\documentclass[accepted]{uai2026} %
\usepackage[american]{babel}
\usepackage{natbib} %
\usepackage{mathtools} %
\usepackage{booktabs} %
\usepackage{tikz} %
\usepackage{pgfplots}
\pgfplotsset{compat=1.18}
\usepackage{amsmath}
\usepackage{amssymb}
\usepackage{subcaption}
\usepackage{url}

\title{Robust Counterfactual Policy Optimisation via\\Nondeterministic Causal Models}

\author[1]{\href{mailto:<jessica.lally@kcl.ac.uk>?Subject=Your UAI 2026 paper}{Jessica Lally}{}}
\author[2]{Milad Kazemi}
\author[1]{Nicola Paoletti}
\author[1]{David Watson}
\author[3]{Sander Beckers}
\affil[1]{%
    King's College London
}
\affil[2]{%
    Durham University
}
\affil[3]{%
    University College London
  }
  
\begin{document}
\maketitle

\begin{abstract}
Counterfactual inference approaches for sequential decision-making typically assume deterministic causal models, where all randomness stems from latent variables. However, Markov Decision Processes (MDPs) are inherently stochastic. We address this by formalising counterfactual policy optimisation under probabilistic nondeterministic causal models, which properly separates latent confounding from irreducible stochasticity, and here propose a first practical optimisation problem for identifying robust counterfactual policies under a sensitivity analysis framework. We validate our approach on a sepsis treatment simulator, where diabetes status acts as a hidden global confounder.
\end{abstract}

\section{Introduction}\label{sec:intro}
Counterfactual inference in sequential decision-making asks: given an observed sequence of states, actions, and outcomes, what would the outcome have been under a different policy? In safety-critical domains, such as healthcare, rerunning experiments under alternative policies is often infeasible or unethical, making counterfactual reasoning crucial for offline policy evaluation.

Existing work applying counterfactual inference to MDPs \citep{oberst2019counterfactual, tsirtsis2021counterfactual, killian2022counterfactually, tsirtsis2024finding, lally2026robustcounterfactualinferencemarkov} assumes all randomness comes from unobserved latent variables, and therefore models MDPs as deterministic structural causal models (SCMs) \citep{pearl_2009}.
However, in practice, randomness can arise from fundamentally different sources: at one extreme, all randomness may arise from unobserved latent variables (the deterministic SCM setting); at the other, all randomness may be irreducible (e.g., from random number generators). Recently, probabilistic nondeterministic causal models (PNSCMs) \citep{beckers2025large, pmlr-v275-beckers25b} have been developed to allow for such inherent stochasticity. More commonly, however, both sources of randomness coexist: health outcomes usually depend both on unobserved latent factors, and on random differences in applying the treatment, or random individual-level genetic variation. Similarly, in the social sciences, variation in the outcome of policies is both due to latent factors and to individual-level variation that is represented as noise at the aggregate level.

In these settings, we propose modelling the environment as a PNSCM where randomness arises from both latent variables and irreducible stochasticity, and bounding the influence of latent variables using a sensitivity-analysis approach similar to \citet{kausik2024offline}. In this paper, we formalise counterfactual inference for MDPs under PNSCMs, and propose optimisation procedures for deriving robust counterfactual policies under confounding. We then evaluate these approaches on the Sepsis MDP \citep{oberst2019counterfactual}, where diabetes acts as an unobserved global confounder, demonstrating that the derived policies are robust when evaluated on the true, fully-specified environment.

\section{Background}
In this section, we introduce the necessary background on Markov decision processes and causal models.

\subsection{Markov Decision Processes}
Markov decision processes (MDPs) model sequential decision-making under uncertainty, and are defined by a tuple $(\mathcal{S}, \mathcal{A}, 
\mathcal{P}_I, \mathcal{P}, \mathcal{R}, \gamma)$ where $\mathcal{S}$ is the state space, $\mathcal{A}$ is the action space, $\mathcal{P}_I$ is the initial state distribution, $\mathcal{P}: \mathcal{S} \times \mathcal{A} \times \mathcal{S} \rightarrow 
[0, 1]$ is the transition kernel, $\mathcal{R}: \mathcal{S} \times 
\mathcal{A} \rightarrow \mathbb{R}$ is a reward function, and $\gamma$ is a discount factor. The goal is to identify an optimal policy $\pi : \mathcal{S} \rightarrow \mathcal{A}$, that optimises the expected total discounted reward. A trajectory $\tau$ under policy $\pi$ is a sequence $\tau = (s_0, a_0, s_1, a_1, \ldots, s_{|\tau|-1}, a_{|\tau|-1}, s_{|\tau|})$ where $s_0 \sim \mathcal{P}_I$, $a_t = \pi(s_t)$, and $s_{t+1} \sim \mathcal{P}(\cdot \mid s_t, a_t)$.

\paragraph{Latent Variables and Confounding}
In many applications, the observed state cannot capture all factors influencing system dynamics, e.g., electronic health records may have missing information on patient demographics or underlying conditions. We consider settings where latent factors $U$ affect transition dynamics, such that the true transitions $P(s' \mid s, a, u)$ differ from the observed marginal $P(s' \mid s, a)$. In this initial paper, we focus on \textit{global confounders}, where the value of $U$ is fixed throughout the trajectory, confounding all transitions \citep{kausik2024offline}.

Since $U$ is unobserved, its effect on transitions cannot be identified from observational data alone. Sensitivity analysis addresses this problem by bounding the degree to which $U$ can influence the transitions. Existing work on sensitivity analysis of confounding in RL primarily focuses on \emph{off-policy evaluation} \citep{bruns2021modelfree, namkoong2020offpolicy, kausik2024offline, brunssmith2023robustfittedqevaluation}: estimating the expected value of a target policy, given a dataset of observed trajectories under an unknown behaviour policy. Here, we adapt this sensitivity framework for \emph{counterfactual policy analysis} \citep{oberst2019counterfactual, tsirtsis2021counterfactual, tsirtsis2024finding, lally2026robustcounterfactualinferencemarkov}: given an observed trajectory, what might have happened under a different policy, and what policy would have been optimal?

\subsection{Structural Causal Models}
Structural causal models (SCMs) \citep{pearl_2009} provide a framework for counterfactual inference. An SCM $\mathcal{C}=(\mathbf{U},\mathbf{V},\mathcal{F},P(\mathbf{U}))$ consists of observed variables $\mathbf{V}$, unobserved variables $\mathbf{U}$  with joint distribution $P(\mathbf{U})$, and structural equations $(X=f_X(\mathbf{pa_X})) \in \mathcal{F}$
that determine the values of each $X \in \mathbf{V}$ as a function of its direct causes -- parents -- $\mathbf{pa}_X$. Counterfactual inference proceeds by estimating $P(\mathbf{U} \mid \mathbf{v})$
given an observation $\mathbf{V}=\mathbf{v}$, performing an intervention that modifies the structural equations, and evaluating the values of observed variables.

\emph{Probabilistic nondeterministic SCMs} (PNSCMs) \citep{pmlr-v275-beckers25b} replace deterministic structural equations $f_X$ with conditional probability distributions $\mathcal{P}_X(\cdot \mid  \mathbf{pa}_X)$, thereby allowing probabilistic causal mechanisms that are not reducible to deterministic mechanisms over unobserved variables. A fully specified MDP together with a policy is an instance of a PNSCM. 
A key feature of counterfactual inference with PNSCMs is \textit{actualised refinement}: given the values of the observed and unobserved variables that produce the actual setting $(\mathbf{u}, \mathbf{v})$, actualised refinement replaces each conditional distribution  $\mathcal{P}_X$ by $\mathcal{P}_X^{(\mathbf{pa}_X, x)}$, which behaves identically to $\mathcal{P}_X$ for all inputs except that it deterministically returns the observed value $x$ when given the observed parents $\mathbf{pa}_X$. This expresses the nondeterministic semantics that {\em all} we learn from the actual setting is that the actual parents resulted in the actual child, and nothing else (see \citep{beckers2025b,pmlr-v275-beckers25b} for more details and motivation.)

\section{Methodology}\label{sec:methodology}
Given an observed trajectory $\tau = (s_0, a_0, \ldots, s_{|\tau|})$, our goal is to identify the optimal policy $\pi$ that maximises the worst-case counterfactual value $V^{\it CF}(0, s_0)$ under uncertainty over the unobserved global confounder $U$:
\begin{equation}
    \begin{aligned}
    V^{\mathit{CF}}(t, s) = \textstyle\max_{\pi} \textstyle\min_{{P}^{\it CF}_t \in \mathcal{P}^{\it CF, \Delta}_t} \\ \mathbb{E}_{s' \sim {P}^{\it CF}_t(\cdot \mid s, \pi(s))} \left[\mathcal{R}(s, \pi(s)) + \gamma \cdot V^{\mathit{CF}}(t+1, s')\right]
    \end{aligned}
\end{equation}
where $\mathcal{P}^{\it CF, \Delta}_t$ is the set of feasible counterfactual transition probabilities at time $t$, whose constraints are defined below. 

\subsection{Sensitivity Constraints}
Similar to the odds-ratio model used in \citep{bruns2021modelfree, brunssmith2023robustfittedqevaluation, kausik2024offline, bennett2024efficient}, we parametrise the influence of $U$ via a sensitivity parameter $\Delta \geq 1$, which constrains how much $P(s' \mid s, a, u)$ can differ from the marginal $P(s' \mid s, a)$:
$$
\frac{1}{\Delta} \leq 
\frac{\text{odds}(P(S'=s'\mid S=s, A=a, U=u))}{\text{odds}(P(S'=s'\mid S=s, A=a))}
\leq \Delta
$$
where $\text{odds}(p) = p/(1-p)$. When $\Delta=1$, this is equivalent to a fully specified PNSCM (i.e., no latent $U$). As $\Delta \rightarrow \infty$, the bounds on $P(S' \mid S, A, U)$ approach $[0, 1]$, representing maximum confounding.
$\Delta$ can be global -- i.e., one value -- or state-action dependent $\Delta(s, a)$ \citep{bennett2024efficient}. 

\subsection{Choosing $\Delta$}
\label{sec: delta}
A key challenge in sensitivity analysis is choosing $\Delta$. Most existing work either delegates this to a domain expert, based on their understanding of the environment dynamics (e.g., \citep{bruns2021modelfree, bennett2024efficient}), or evaluates over a range from $\Delta=1$ (no confounding) to some large value (strong confounding), assessing how the estimated value function changes. If the target policy consistently outperforms the observed policy, this demonstrates robustness. Without domain knowledge, a more principled approach is to calibrate $\Delta$ from the observed state variables \citep{brunssmith2023robustfittedqevaluation} by hiding each variable in turn and computing the resulting odds ratio between the observed and marginalised transitions. Under the assumption that no unobserved variable influences transitions more than the strongest observed variable, this provides a justifiable upper bound on $\Delta(s, a)$. Formally, for each variable $i$, let $s^{-i}$ denote the state omitting variable $i$, and let $n_i$ denote the number of distinct values variable $i$ can take. The marginalised transition distribution when variable $i$ is hidden is:
\begin{equation}
\begin{aligned}
    P(S_{t+1} = s' \mid S^{-i}_t = s^{-i}, A_t = a) = 
    \sum_{j=0}^{n_i - 1} w_i(j \mid s^{-i}) \cdot \\
    P(s' \mid s^{-i}, S^i = j, a)
\end{aligned}
\end{equation}
where $w_i(j \mid s^{-i}) = P(S^i = j \mid s^{-i})$ is the conditional probability that variable $i$ takes value $j$ given the projected state $s^{-i}$, estimated from the stationary distribution of the MDP. The sensitivity $\Delta(s, a)$ can be computed as:
\begin{equation*}
    \Delta_i(s, a) = \max_{\substack{s' \in \mathcal{S} \\ P(s' \mid s, a) > 0}} 
    \max\left(r_i(s, a, s'),\ r_i(s, a, s')^{-1}\right)
\end{equation*}
where $r_i(s, a, s') = \dfrac{\text{odds}\left(P(S_{t+1} = s' \mid s, a)\right)}
    {\text{odds}\left(P(S_{t+1} = s' \mid s^{-i}, a)\right)}$.
    
The overall sensitivity for each state-action pair is $\Delta(s, a) = \max_i \Delta_i(s, a).$
Additionally, a scaling factor $\Gamma \geq 0$ can be applied to calibrate the sensitivity, giving a final sensitivity estimate of $\Gamma \cdot \Delta(s, a)$ \citep{mcclean2025calibratedsensitivitymodels}. This provides an interpretable parameter to assess how counterfactual estimates and policies vary under stronger $(\Gamma > 1)$ or weaker $(\Gamma < 1)$ confounding than the proxy $\Delta(s, a)$ suggests. 

\subsection{Actualised Refinement}
\label{sec: actualised refinement}
Given an observed transition $(s_t, a_t, s_{t+1})$, if we were to take the same action $a_t$ in $s_t$ in some counterfactual world, the  next state would be the same $s_{t+1}$ as well. The reason is that no matter the unobserved value $U=u$, it remains constant across worlds. Still, given that we do not know the actual value $U=u$, the counterfactual transition probability must reflect this posterior uncertainty: %
with probability $P(U \mid \tau)$, $U=u$ generated the trajectory and the transition to $s_{t+1}$ is certain; otherwise, with probability $1-P(U \mid \tau)$, the counterfactual probability equals the estimated probability $\tilde{P}(s' \mid s_t, a_t, u)$ as chosen by the optimisation. As a result, the counterfactual transition probability is:
\begin{equation*}
    P^{CF}_t(s'|s,a,u) = \begin{cases}
        \begin{aligned}
            &P(U=u|\tau)\cdot\mathbf{1}[s'=s_{t+1}] \\
            &\quad + (1-P(U=u|\tau))\cdot\tilde{P}(s'|s,a,u)
        \end{aligned} \\
        \qquad\qquad\qquad\qquad \text{if } (s,a) = (s_t, a_t) \\[4pt]
        \tilde{P}(s'|s,a,u) \quad \text{otherwise}
    \end{cases}
\end{equation*}
(One exception is when $\Delta=1$: this corresponds to a setting with no confounder $U$, so $P^{CF}_t(s_{t+1}\mid s_t,a_t) = 1$).

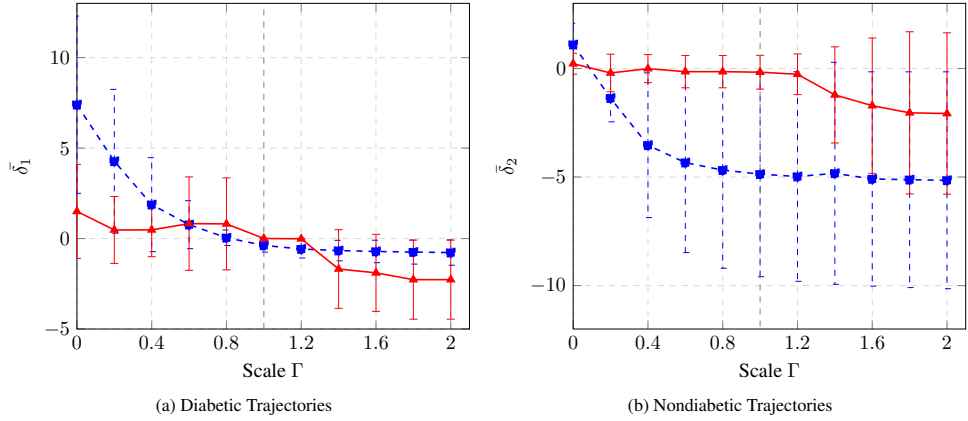
\begin{figure*}
\centering
\resizebox{1.0\textwidth}{!}{%
    \begin{tikzpicture}
    \begin{axis}[
    width=\textwidth,
    hide axis,
    xmin=0, xmax=1, ymin=0, ymax=1,
    height=0.5cm,
    scale only axis,
    legend style={at={(0.5,0.5)}, anchor=center, legend columns=3, font=\small},
]
\addplot[color=blue, mark=square*, thick, dashed] coordinates {(10,10) (11,11)};
\addlegendentry{Mean Difference $\bar{\delta_1} = V^{\mathit{CF}}_{\Delta, \Gamma}(0, s_0) - V^{\mathit{CF}}_{\Delta^*}(0, s_0)$}
\addplot[color=red, mark=triangle*, thick, solid] coordinates {(10,10) (11,11)};
\addlegendentry{Mean Difference $\bar{\delta_2} = V^{\mathit{CF}, \pi^*_{\Delta, \Gamma}}_{\it full}(0, s_0) - V^{\mathit{CF}, \pi^*_{\Delta*}}_{\it full}(0, s_0)$}
\addplot[dashed, gray, forget plot] coordinates {(1, 10) (1, 10)};
\addlegendimage{dashed, gray, ybar, bar width=0.1pt}
\addlegendentry{No scaling ($\Gamma = 1$)}
    \end{axis}
    \end{tikzpicture}}\\
    \vspace{0.1cm}
\resizebox{0.75\textwidth}{!}{%
    \begin{minipage}{0.49\textwidth}
    \begin{tikzpicture}
    \begin{axis}[
        xlabel={Scale $\Gamma$},
        ylabel={$\bar{\delta_1}$},
        xmin=0, xmax=2.1,
        ymin=-5, ymax=13,
        xtick={0.0, 0.4, 0.8, 1.2, 1.6, 2.0},
        legend style={font=\small, at={(0.5,-0.25)}, anchor=north east, legend columns=1},
        grid=major, grid style={dashed, gray!30},
        error bars/y dir=both, error bars/y explicit,
    ]
    \addplot[
        color=blue,
        mark=square*,
        thick,
        dashed,
        error bars/y dir=both,
        error bars/y explicit,
        forget plot
    ] coordinates {
        (0.0, 7.4) +- (0, 4.9)
        (0.2, 4.27087)  +- (0, 3.975189)
        (0.4, 1.87491)  +- (0, 2.596275)
        (0.6, 0.76724)  +- (0, 1.325421)
        (0.8, 0.04453)  +- (0, 0.428129)
        (1.0, -0.37713) +- (0, 0.373321)
        (1.2, -0.57863) +- (0, 0.49655)
        (1.4, -0.66461) +- (0, 0.56739)
        (1.6, -0.71301) +- (0, 0.620743)
        (1.8, -0.74762) +- (0, 0.66423)
        (2.0, -0.77529) +- (0, 0.699423)
    };

    \addplot[
        color=red,
        mark=triangle*,
        thick,
        solid,
        error bars/y dir=both,
        error bars/y explicit,
        forget plot
    ] coordinates {
        (0, 1.5) +- (0, 2.6)
        (0.2, 0.47395)  +- (0, 1.847827)
        (0.4, 0.47793)  +- (0, 1.480981)
        (0.6, 0.82511)  +- (0, 2.582222)
        (0.8, 0.80978)  +- (0, 2.54291)
        (1.0, 0.00389)  +- (0, 0.006479)
        (1.2, -0.01548) +- (0, 0.031756)
        (1.4, -1.68227) +- (0, 2.174931)
        (1.6, -1.89594) +- (0, 2.129934)
        (1.8, -2.27209) +- (0, 2.185597)
        (2.0, -2.27209) +- (0, 2.185597)
    };
    \addplot[dashed, gray] coordinates {(1, -5) (1, 13)};

    \end{axis}
    \end{tikzpicture}
    \subcaption{Diabetic Trajectories}
    \label{fig:cf_value_gamma_diabetic}
    \end{minipage}
    \begin{minipage}{0.49\textwidth}
    \begin{tikzpicture}
    \begin{axis}[
        xlabel={Scale $\Gamma$},
        ylabel={$\bar{\delta}_2$},
        xmin=0, xmax=2.1,
        ymin=-12, ymax=3,
        xtick={0.0, 0.4, 0.8, 1.2, 1.6, 2.0},
        legend style={font=\small, at={(0.5,-0.25)}, anchor=north, legend columns=1},
        grid=major, grid style={dashed, gray!30},
        error bars/y dir=both, error bars/y explicit,
    ]
    \addplot[
        color=blue,
        mark=square*,
        thick,
        dashed,
        error bars/y dir=both,
        error bars/y explicit,,
        forget plot
    ] coordinates {
        (0, 1.1) +- (0, 0.98)
        (0.2, -1.36702)  +- (0, 1.090521)
        (0.4, -3.53445)  +- (0, 3.331191)
        (0.6, -4.33579)  +- (0, 4.143906)
        (0.8, -4.68205)  +- (0, 4.521668)
        (1.0, -4.86608)  +- (0, 4.722042)
        (1.2, -4.97700)  +- (0, 4.826788)
        (1.4, -4.82879)  +- (0, 5.113169)
        (1.6, -5.08941)  +- (0, 4.935477)
        (1.8, -5.12297)  +- (0, 4.968601)
        (2.0, -5.14981)  +- (0, 4.99518)
    };

    \addplot[
        color=red,
        mark=triangle*,
        thick,
        solid,
        error bars/y dir=both,
        error bars/y explicit,
        forget plot
    ] coordinates {
       (0, 0.22) +- (0, 0.48)
        (0.2, -0.20706)  +- (0, 0.864947)
        (0.4, -0.00836)  +- (0, 0.646935)
        (0.6, -0.14666)  +- (0, 0.743137)
        (0.8, -0.14673)  +- (0, 0.74313)
        (1.0, -0.17137)  +- (0, 0.778541)
        (1.2, -0.26608)  +- (0, 0.936205)
        (1.4, -1.21705)  +- (0, 2.213831)
        (1.6, -1.71244)  +- (0, 3.115894)
        (1.8, -2.04286)  +- (0, 3.734312)
        (2.0, -2.07425)  +- (0, 3.716219)
    };
    \addplot[dashed, gray] coordinates {(1, -13) (1, 3)};
    \end{axis}
    \end{tikzpicture}
    \subcaption{Nondiabetic Trajectories}
    \label{fig:cf_value_gamma_nondiabetic}
    \end{minipage}
}
\caption{Mean difference between counterfactual policy values and the oracle (under true diabetes sensitivity $\Delta^*$), averaged across $10$ randomly sampled suboptimal trajectories.}
\end{figure*}
\subsection{Optimisation Procedures}
Due to the coupling of $P(S' \mid S, A, U)$ between all state-action pairs $(s, a)$ and all time steps $t$, identifying globally optimal solutions is challenging on all but toy examples. In this work, we therefore consider a {\bf decoupled approach}: we simplify the optimisation problem by removing the time-homogeneity assumption and the coupling between state-action pairs $(s, a)$, treating each $(t, s, a)$ triple independently. For each $(t,s)$ we solve:
\begin{equation*}
\begin{aligned}
    &V^{\it CF}(t,s) = \max_a \big[ R(s,a) + \gamma \cdot \textstyle\min_{\substack{\{\tilde{P}(s_{t+1} \mid s_t,a_t,u)\}_{u} \\ \{\tilde{P}(s'\mid s,a,u)\}_{u,s'}}} \\&\sum_u P(U=u|\tau) \sum_{s'} P^{CF}_t(s'|s,a,u) \cdot V^{\it CF}(t+1,s') \big]
\end{aligned}
\end{equation*}
where the posterior $P(U=u \mid \tau)$ is:
\begin{equation*}
    P(U=u \mid \tau) = \frac{P(U=u) \prod_t \tilde{P}(s_{t+1} \mid s_t,a_t,u)}{\sum_{u'} P(U=u')\prod_t \tilde{P}(s_{t+1}\mid s_t,a_t,u')}
\end{equation*}
Since $P(U \mid \tau)$ is a continuous function of the adversary's choice of transition probabilities, we compute its feasible range by setting each observed transition $\tilde{P}(s_{t+1} \mid s_t, a_t, u)$ to its extremes and treat it as an optimisation variable. The adversary therefore jointly optimises over $P(U \mid \tau)$ and $\{\tilde{P}(\cdot|s,a,u)\}_u$ at each $(t,s,a)$ block, subject to:
\begin{equation*}
\begin{aligned}
    \textstyle\sum_u P(U=u) \cdot \tilde{P}(s'|s,a,u) = P(s'|s,a), \quad \forall s' \\
    \textstyle\sum_{s'} \tilde{P}(s'|s,a,u) = 1, \quad \forall u \\
    \tfrac{1}{\Delta} \cdot P(s'|s,a) \leq \tilde{P}(s'|s,a,u) \leq \Delta \cdot P(s'|s,a), \quad \forall s', u
\end{aligned}
\end{equation*}
Note that at the observed transition $(s,a) = (s_t, a_t)$, $P(U=u \mid \tau)$ appears both as a posterior weight and inside $P^{CF}_t$, creating an adversarial tradeoff that makes the optimisation non-convex; we use Gurobi to guarantee a globally optimal solution.
The decoupling makes the problem tractable but produces a conservative lower bound on $V^{\mathit CF}(0, s_0)$: the decoupled adversary has strictly more freedom than the true coupled adversary (which must use the same $\tilde{P}(s' \mid s, a, u)$ and $P(U \mid \tau)$ across all $(t,s,a)$ blocks), and since greater adversarial freedom can only decrease the value, this guarantees a valid lower bound.

\paragraph{Incorporating Time-homogeneity}
To obtain a tighter bound, we could incorporate time-homogeneity by requiring a single $P(s' \mid s, a, u)$ for each $(s, a)$ pair across all timesteps. This creates a trilinear coupling between the posterior $P(U \mid \tau)$, transition probabilities $P(s' \mid s, a, u)$, and value function, which can be solved using gradient descent. Since gradient descent finds only local minima, we cannot guarantee we will identify the true pessimistic value function. However, the decoupled and gradient descent solutions will together bound the true pessimistic value function.

\section{Sepsis Simulation}\label{sec:evaluation}
We demonstrate our approach on a sepsis treatment simulator \citep{oberst2019counterfactual}. Each state consists of four vital signs (heart rate, blood pressure, oxygen concentration, and glucose levels), categorised as low, normal, or high. At each step, three treatments can be toggled on or off. Rewards range from $-10$ (death) to $10$ (discharge) based on the number of out-of-range vital signs. The full transition matrix is $P(S' \mid S, A, U)$, where $U \in \{0, 1\}$ indicates whether the patient is diabetic, which acts as a global latent confounder. In this population, $20\%$ of patients are diabetic.

We evaluate our approach on the marginalised transition matrix (where diabetes is hidden), given observed trajectories from a suboptimal policy (which acts optimally with probability $0.5$, and randomly otherwise). We use the maximum sensitivity of the observed vital signs $\Delta(s, a)$ (see Section \ref{sec: delta}) as a proxy for the unobserved diabetes sensitivity, and an optional scaling parameter $\Gamma$. For each trajectory, we derive the optimal pessimistic counterfactual policy $\pi^*_\Delta$ that maximises the worst-case counterfactual value function $V^{\it CF}_{\Delta, \Gamma}(0, s_0)$, and evaluate its performance on the fully-observed MDP (where the diabetes state is known) to obtain $V^{\it CF, \pi^*_\Delta}_{\it full}(0, s_0)$. We evaluate our approach along two axes: first, whether $\pi^*_\Delta$ improves on the observation; and second, how closely our estimated $V^{\it CF}_{\Delta, \Gamma}(0, s_0)$ and policies match those obtained under the true diabetes sensitivity $\Delta^*$.

\paragraph{Improvement over Observation} Table \ref{tab:cf_value_observed} shows that our approach can identify counterfactual policies that robustly improve on the observed suboptimal behaviour when evaluated on the fully-observed MDP. The estimated values $V^{\it CF}_{\Delta, \Gamma}(0, s_0)$ underestimate this improvement, particularly for nondiabetic trajectories as the $\Delta(s, a)$ proxy overestimates the confounding effect. This is expected, as the marginalised transition matrix already closely reflects nondiabetic dynamics (since $P(U=0)=0.8)$.

\begin{table}[h]
\centering
\caption{Mean difference between counterfactual policy values and observed discounted return $G_0$ ($\gamma = 0.9$), across $10$ suboptimal diabetic and nondiabetic trajectories. \textit{Estimated} reports the mean difference $V^{\it CF}_{\Delta, \Gamma}(0, s_0)-G_0$; \textit{Actual} reports the mean difference $V^{\it CF, \pi^*_\Delta}_{\it full}(0, s_0)-G_0$.}
\label{tab:cf_value_observed}
\setlength{\tabcolsep}{6pt}
\begin{tabular}{cc cc}
\toprule
\multicolumn{2}{c}{\textbf{Diabetic}} 
& \multicolumn{2}{c}{\textbf{Nondiabetic}} \\
\cmidrule(lr){1-2} \cmidrule(lr){3-4}
Estimated & Actual & Estimated & Actual \\
\midrule
$3.4 \pm 10.9$ & $7.4 \pm 10.7$ & $-0.50 \pm 9.1$ & $5.0 \pm 7.4$ \\
\bottomrule
\end{tabular}
\end{table}

\paragraph{Comparison to Diabetes Sensitivity} Figures \ref{fig:cf_value_gamma_diabetic} and \ref{fig:cf_value_gamma_nondiabetic} assess how well our proxy $\Delta(s, a)$ approximates the true diabetes sensitivity $\Delta^*$ (the oracle). Essentially, $\bar{\delta_1}$ (the blue curve) reflects how close our estimated counterfactual values are to the oracle, and $\bar{\delta_2}$ (the red curve) reflects how close the policies derived from our proxy $\Delta(s, a)$ perform relative to the oracle in practice. The oracle represents the most accurate estimates achievable by our robust pessimistic approach, if we knew the true sensitivity $\Delta^*$. Since $\Delta(s, a) \geq 1$, we clamp $\Gamma \cdot \Delta(s, a)$ to a minimum of $1$ (so $\Gamma=0$ corresponds to a global sensitivity $\Delta=1$, i.e., no confounding). At $\Gamma=0$, $\bar{\delta_1} > 0$ (and is much greater than $0$ for diabetic trajectories), confirming that ignoring confounding leads to inaccurate estimates. For diabetic trajectories, $\Gamma=1.0$ minimises both $\bar{\delta_1}$ and $\bar{\delta_2}$, demonstrating that the unscaled proxy reliably estimates the true diabetes sensitivity. For nondiabetic trajectories, $\bar{\delta_1}$ is strongly negative at $\Gamma=1$, and approaches $0$ as we reduce $\Gamma$, showing our estimates are conservative. This is expected, as the marginalised transition matrix already closely reflects nondiabetic dynamics, so the true confounding effect for nondiabetics is weaker than the proxy suggests. However, crucially, $\bar{\delta_2}$ remains close to $0$ across $0 < \Gamma \leq 1$, suggesting that despite the conservative value estimates, we recover policies close to the oracle.

\section{Conclusion}
In this paper, we introduced a framework for counterfactual policy optimisation under global confounding and demonstrated its robustness on the Sepsis MDP. Future work will explore extending this framework to settings with different types of latent variables (e.g., memoryless confounders, and confounders that persist over a subset of the time steps) and evaluating the gradient descent-based optimisation method to obtain more accurate counterfactual value estimates.

\bibliography{uai2026-template}

\begin{thebibliography}{15}
\providecommand{\natexlab}[1]{#1}
\providecommand{\url}[1]{\texttt{#1}}
\expandafter\ifx\csname urlstyle\endcsname\relax
  \providecommand{\doi}[1]{doi: #1}\else
  \providecommand{\doi}{doi: \begingroup \urlstyle{rm}\Url}\fi

\bibitem[Beckers(2025{\natexlab{a}})]{beckers2025b}
Sander Beckers.
\newblock Causal counterfactuals reconsidered.
\newblock \emph{arXiv preprint arXiv:2512.12804}, 2025{\natexlab{a}}.

\bibitem[Beckers(2025{\natexlab{b}})]{pmlr-v275-beckers25b}
Sander Beckers.
\newblock Nondeterministic causal models.
\newblock In Biwei Huang and Mathias Drton, editors, \emph{Proceedings of the
  Fourth Conference on Causal Learning and Reasoning}, volume 275 of
  \emph{Proceedings of Machine Learning Research}, pages 1532--1554. PMLR,
  07--09 May 2025{\natexlab{b}}.
\newblock URL \url{https://proceedings.mlr.press/v275/beckers25b.html}.

\bibitem[Beckers(2026)]{beckers2025large}
Sander Beckers.
\newblock Large language models as nondeterministic causal models.
\newblock In \emph{Proceedings of the 23rd International Conference on
  Principles of Knowledge Representation and Reasoning}, 2026.

\bibitem[Bennett et~al.(2024)Bennett, Kallus, Oprescu, Sun, and
  Wang]{bennett2024efficient}
Andrew Bennett, Nathan Kallus, Miruna Oprescu, Wen Sun, and Kaiwen Wang.
\newblock Efficient and sharp off-policy evaluation in robust markov decision
  processes.
\newblock In A.~Globerson, L.~Mackey, D.~Belgrave, A.~Fan, U.~Paquet,
  J.~Tomczak, and C.~Zhang, editors, \emph{Advances in Neural Information
  Processing Systems}, volume~37, pages 112962--113000. Curran Associates,
  Inc., 2024.
\newblock \doi{10.52202/079017-3590}.
\newblock URL \url{https://doi.org/10.52202/079017-3590}.

\bibitem[Bruns-Smith and Zhou(2023)]{brunssmith2023robustfittedqevaluation}
David Bruns-Smith and Angela Zhou.
\newblock Robust fitted-q-evaluation and iteration under sequentially exogenous
  unobserved confounders, 2023.
\newblock URL \url{https://arxiv.org/abs/2302.00662}.

\bibitem[Bruns-Smith(2021)]{bruns2021modelfree}
David~A Bruns-Smith.
\newblock Model-free and model-based policy evaluation when causality is
  uncertain.
\newblock In Marina Meila and Tong Zhang, editors, \emph{Proceedings of the
  38th International Conference on Machine Learning}, volume 139 of
  \emph{Proceedings of Machine Learning Research}, pages 1116--1126. PMLR,
  18--24 Jul 2021.
\newblock URL \url{https://proceedings.mlr.press/v139/bruns-smith21a.html}.

\bibitem[Kausik et~al.(2024)Kausik, Lu, Tan, Makar, Wang, and
  Tewari]{kausik2024offline}
Chinmaya Kausik, Yangyi Lu, Kevin Tan, Maggie Makar, Yixin Wang, and Ambuj
  Tewari.
\newblock Offline policy evaluation and optimization under confounding.
\newblock In Sanjoy Dasgupta, Stephan Mandt, and Yingzhen Li, editors,
  \emph{Proceedings of The 27th International Conference on Artificial
  Intelligence and Statistics}, volume 238 of \emph{Proceedings of Machine
  Learning Research}, pages 1459--1467. PMLR, 02--04 May 2024.
\newblock URL \url{https://proceedings.mlr.press/v238/kausik24a.html}.

\bibitem[Killian et~al.(2022)Killian, Ghassemi, and
  Joshi]{killian2022counterfactually}
Taylor~W Killian, Marzyeh Ghassemi, and Shalmali Joshi.
\newblock Counterfactually guided policy transfer in clinical settings.
\newblock In \emph{Conference on Health, Inference, and Learning}, pages 5--31.
  PMLR, 2022.

\bibitem[Lally et~al.(2026)Lally, Kazemi, and
  Paoletti]{lally2026robustcounterfactualinferencemarkov}
Jessica Lally, Milad Kazemi, and Nicola Paoletti.
\newblock Robust counterfactual inference in markov decision processes.
\newblock AAMAS '26, page 1527–1535, Richland, SC, 2026. International
  Foundation for Autonomous Agents and Multiagent Systems.
\newblock ISBN 9798400723179.
\newblock \doi{10.65109/TXUQ4572}.
\newblock URL \url{https://doi.org/10.65109/TXUQ4572}.

\bibitem[McClean et~al.(2025)McClean, Branson, and
  Kennedy]{mcclean2025calibratedsensitivitymodels}
Alec McClean, Zach Branson, and Edward~H. Kennedy.
\newblock Calibrated sensitivity models, 2025.
\newblock URL \url{https://arxiv.org/abs/2405.08738}.

\bibitem[Namkoong et~al.(2020)Namkoong, Keramati, Yadlowsky, and
  Brunskill]{namkoong2020offpolicy}
Hongseok Namkoong, Ramtin Keramati, Steve Yadlowsky, and Emma Brunskill.
\newblock Off-policy policy evaluation for sequential decisions under
  unobserved confounding.
\newblock In H.~Larochelle, M.~Ranzato, R.~Hadsell, M.F. Balcan, and H.~Lin,
  editors, \emph{Advances in Neural Information Processing Systems}, volume~33,
  pages 18819--18831. Curran Associates, Inc., 2020.
\newblock URL
  \url{https://proceedings.neurips.cc/paper_files/paper/2020/file/da21bae82c02d1e2b8168d57cd3fbab7-Paper.pdf}.

\bibitem[Oberst and Sontag(2019)]{oberst2019counterfactual}
Michael Oberst and David Sontag.
\newblock Counterfactual off-policy evaluation with gumbel-max structural
  causal models.
\newblock In \emph{International Conference on Machine Learning}, pages
  4881--4890. PMLR, 2019.

\bibitem[Pearl(2009)]{pearl_2009}
Judea Pearl.
\newblock \emph{Causality}.
\newblock Cambridge University Press, 2$^\textnormal{nd}$ edition, 2009.
\newblock \doi{10.1017/CBO9780511803161}.

\bibitem[Tsirtsis and Rodriguez(2024)]{tsirtsis2024finding}
Stratis Tsirtsis and Manuel Rodriguez.
\newblock Finding counterfactually optimal action sequences in continuous state
  spaces.
\newblock \emph{Advances in Neural Information Processing Systems}, 36, 2024.

\bibitem[Tsirtsis et~al.(2021)Tsirtsis, De, and
  Rodriguez]{tsirtsis2021counterfactual}
Stratis Tsirtsis, Abir De, and Manuel Rodriguez.
\newblock Counterfactual explanations in sequential decision making under
  uncertainty.
\newblock \emph{Advances in Neural Information Processing Systems},
  34:\penalty0 30127--30139, 2021.

\end{thebibliography}

\end{document}